%% file: ijcai24.tex
\documentclass{article}
\usepackage{ijcai24}

\usepackage{times}
\usepackage{soul}
\usepackage{url}
\usepackage[hidelinks]{hyperref}
\usepackage[utf8]{inputenc}
\usepackage[small]{caption}
\usepackage{graphicx}
\usepackage{amssymb,amsmath,amsthm, amsfonts}
\usepackage{booktabs}
\usepackage{algorithm}
\usepackage{algorithmic}
\usepackage[switch]{lineno}

\usepackage{enumitem}
\usepackage{mathtools}
\usepackage{caption}
\usepackage{subcaption}
\usepackage{threeparttable}
\usepackage{booktabs}
\setlist[description]{leftmargin=*}
\usepackage{amssymb,amsmath,amsthm, amsfonts}

\usepackage{xspace}
\usepackage{newfloat}
\usepackage{listings}

\newtheorem{theorem}{Theorem}
\usepackage{newfloat}
\usepackage{listings}

\newcommand{\wrf}[1]{{#1}}

\newcommand{\opauc}{opAUC\xspace}
\newcommand{\recall}{recall\xspace}
\newcommand{\ours}{TBL}

\newcommand{\result}[2]{${#1}_{\pm#2}$}
\newcommand{\resultf}[2]{$\mathbf{#1}_{\pm#2}$}

\newtheorem{remark}{Remark}
\newtheorem{definition}[theorem]{Definition}

\title{Ultra-imbalanced classification guided by statistical information}

\author{
Yin Jin$^{1,2}$
\and
Ningtao Wang$^2$\And
Ruofan Wu$^2$\And
Pengfei Shi$^2$\And
Xing fu$^2$\And\\
Weiqiang Wang$^2$
\affiliations
$^1$Center for data Science, Zhejiang university, Hangzhou, China\\
$^2$Tiansuan Lab, Ant group, Hangzhou, China\\
\emails
jin$\_$yin@zju.edu.cn, ningtao.nt@antgroup.com}

\begin{document}

\maketitle

\begin{abstract}
    Imbalanced data are frequently encountered in real-world classification tasks. Previous works on imbalanced learning mostly focused on learning with a minority class of few samples. However, the notion of imbalance also applies to cases where the minority class contains abundant samples, which is usually the case for industrial applications like fraud detection in the area of financial risk management. In this paper, we take a population-level approach to imbalanced learning by proposing a new formulation called \emph{ultra-imbalanced classification} (UIC). Under UIC, loss functions behave differently even if infinite amount of training samples are available. To understand the intrinsic difficulty of UIC problems, we borrow ideas from information theory and establish a framework to compare different loss functions through the lens of statistical information. A novel learning objective termed Tunable Boosting Loss (\ours) is developed which is provably resistant against data imbalance under UIC, as well as being empirically efficient verified by extensive experimental studies on both public and industrial datasets.
\end{abstract}

\section{Introduction}
\label{introduction}
\subsection{Motivations and contributions}
\begin{figure}
\centering
\includegraphics[width=0.8\columnwidth]{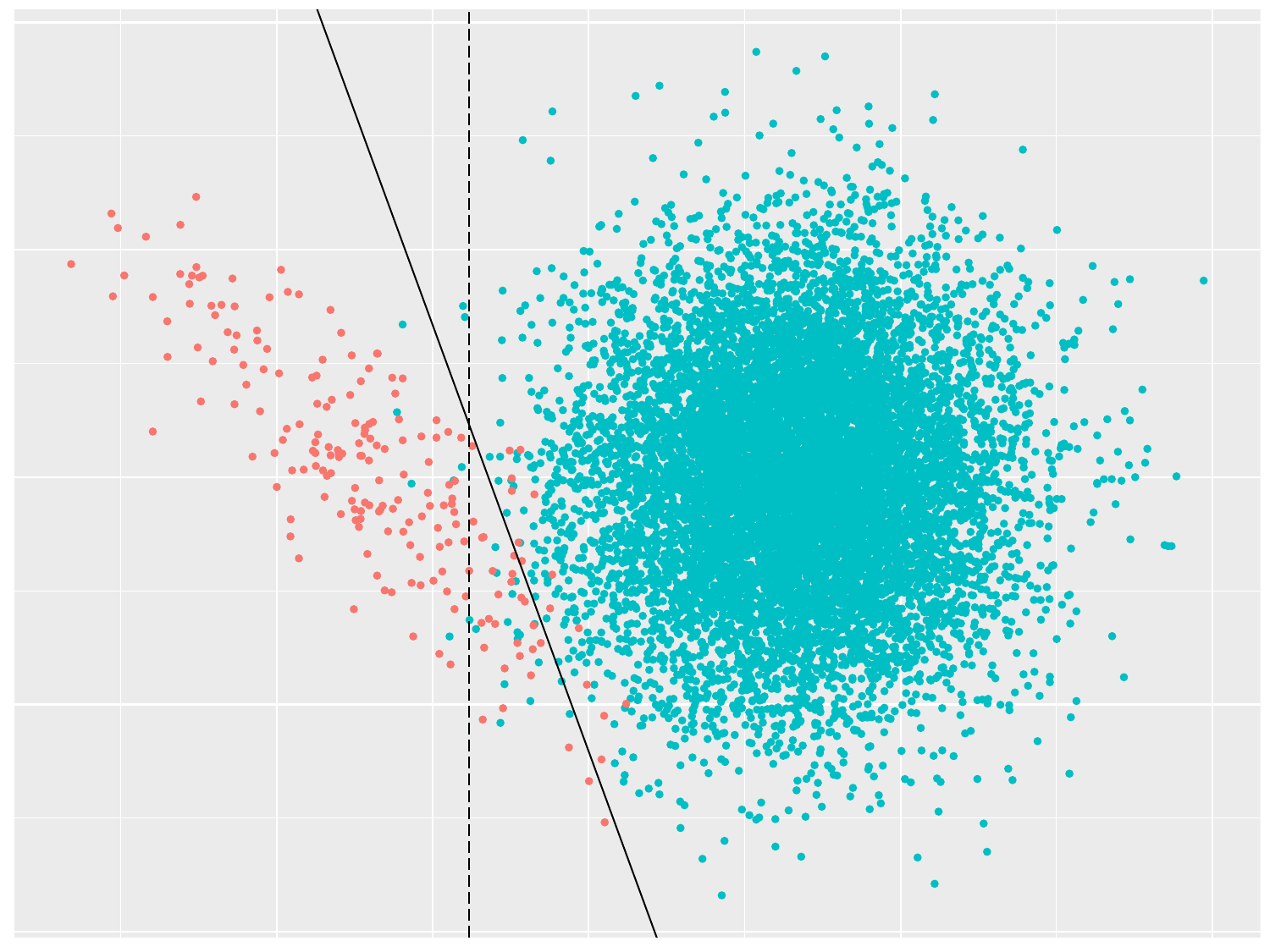}
\caption{Linear classifier learned by different losses on two normal clusters. The ratio of minority samples (red) to majority samples (cyan) is 1:1000. Dashed line: linear classifier learned by cross entropy; Solid line: linear classifier learned by exponential loss.}
\label{fig: introduction}
\end{figure}

Extremely imbalanced training environment dominates real-world learning tasks, such as object detection\cite{tan2020equalization,zhang2021deep}, network intrusion detection \cite{cieslak2006combating} and fraud detection \cite{brennan2012comprehensive}. For example, in a fraud detection task, the ratio of fraud cases can be as low as 1:$10^6$ \cite{foster2004variable}. Training on extremely imbalanced datasets can lead to poor generalization performance due to the large variance brought by the under-presented minority class \cite{wei2022open}. However, challenges still exist even if we have abundant samples from the minority. \wrf{Specifically, classifiers learned via different loss functions behave differently. We present a pictorial illustration in Figure \ref{fig: introduction}, where the data are generated from two normally distributed clusters, with $200$ minority class sample and $200,000$ majority class sample. We plot the decision boundary of linear classifiers learned under cross entropy loss and exponential loss. Despite the fact that the number of minority samples suffices for learning a linear classifier, we observe an intriguing phenomenon that classifier learned under the cross entropy ignores the variance information of the minority class which was captured by the one learned under the exponential loss.}
\wrf{Meanwhile, considerable effort has been made toward designing better loss functions that fit better to the imbalanced regime than standard choices like the cross entropy loss \cite{lin2017focal,ben2020asymmetric,li2019gradient,leng2022polyloss}. Nonetheless, empirical evidences \cite{cao2019learning} suggested that most of such designs occasionally fail in classification scenarios. It is therefore of interest to develop principled frameworks of comparing different loss functions under the imbalanced learning setup. }
 
On the theory side, recent developments on imbalanced classification \cite{kini2021label,zhai2022understanding} mostly focus on establishing theoretical guarantees on \emph{separable data} with a few samples from the minority class using overparameterized models. While such analyses have a nice connection to optimization and modern learning theory, the assumption might not fit in reality. For example, in the area of financial risk management (FRM), the imbalance of training data is sometimes expressed in the sense of \emph{relative rarity} with a potentially large number of minority samples. Under such setups, the separability assumption is unlikely to hold. 
 
\wrf{To address the aforementioned challenges, we take a \emph{population-level perspective} and introduce the concept of \emph{ultra-imbalanced classification} (UIC) as an alternative formulation for imbalanced classification, which means that the prior probability of a sample belonging to the minority class limits to zero. Under the UIC setup, we draw insights from information theory and develop a principled framework for comparing different loss functions inspired from the idea of statistical information \cite{degroot1962uncertainty}. A thorough analysis is conducted regarding the behavior of commonly used loss functions, as well as losses tailor-made for imbalanced problems, showing that learning objectives such as focal loss \cite{lin2017focal} and polyloss \cite{leng2022polyloss} do not provide solid improvement over the cross entropy loss. We summarize our contribution as follows:
\begin{itemize}
    \item We introduce the UIC formulation as a new paradigm for studying imbalanced learning problems. Under the UIC setup, we construct simple cases where the cross entropy objective becomes provably sub-optimal. We also establish the optimality of the recently proposed alpha loss \cite{Sypherd2022tunable} under certain conditions. 
    \item We propose a new framework for comparing different loss functions under UIC. The framework utilizes the concept of statistical information with respect to certain losses and use the decaying rate of the corresponding $f$-function as a measure of resistance against imbalance. As a consequence, we present a systematic study regarding commonly used learning objectives as well as some recently proposed variants under imbalanced learning setup, showing that none of the variants provide solid improvements over the cross entropy objective.
    \item We propose a novel learning objective that is based on a denoising modification of alpha-loss that provably dominates cross entropy under the proposed comparison framework under UIC. Extensive empirical evaluations are conducted to verify the practical efficacy of the proposed objective over both public datasets and two industrial datasets.
\end{itemize}
}

\subsection{Related literatures}
\label{sec:headings}

\textbf{Infinitely imbalance:} 
\cite{owen2007infinitely} discussed the setting where the minority class has a finite sample size and the size of the majority class grows without bound. In that case, the coefficient vector of the logistic regression approaches a useful limit. The setting resembles our ultra-imbalance setting and we have seen similar results in \cite{bach2006considering}. However, we extend our analysis to more general loss functions and introduce the framework of statistical information to help characterize their different behavior under ultra-imbalance.

\noindent
\textbf{Reweighting by class:} To tackle imbalanced data, reweighting samples simply by adjusting the class-wise margin is an intuitive scheme, such as logit adjustment \cite{menon2020long} and  variants \cite{cao2019learning,kini2021label}. This kind of method could be integrated into any loss design, and we delay the discussion in section 2.3. Another common way to address class imbalance is to upweight the minority class by a constant factor, commonly set by inverse class frequency \cite{huang2016learning} or a smoothed version of the inverse square root of class frequency \cite{mikolov2013distributed}. \cite{cui2019class} proposed a weighted factor based on the effective number of samples and practiced better than the trivial choice. But overall, compared to logit adjustment and its variants which uses prior label distribution, the effect of these methods is less remarkable.

\noindent
\textbf{Reweighting by Classification difficulty:}
Many loss functions designed for imbalance classification reweight the samples by their difficulty of classification, which include Focal loss \cite{lin2017focal,ben2020asymmetric}, Equalized loss variants \cite{tan2021equalization,li2022equalized}, Poly loss \cite{leng2022polyloss}, gradient harmonized detector \cite{li2019gradient}.
They share the same motivation of balancing the gradient contribution of different class, since it is hypothesised that the generalization performance could be enhanced by the balance of gradient contribution among different classes \cite{tan2020equalization}. The hypothesis is theoretically supported by \cite{wang2021importance} with the assumption of overparameterized network and separable data. However, empirically their performance in many datasets do not match simple margin-adjusted methods \cite{ye2020identifying}. 

\section{Ultra-imbalance and statistical information}
\subsection{The UIC formulation}

Consider the case of binary classification. We assume $Y\in \{0,1\}$ and $Y = 1$ represents the minority class, with predictors $X \in \mathcal{X}\subset \mathbb{R}^d$. The goal is to discriminate the distribution of $X|Y=1$ and $X|Y=0$ denoted by $P$ and $Q$. We model and estimate the observation-conditional density $\eta(x)=P(Y=1|X=x)$, which gives us a Bayes classifier. We set the prior probability $\pi={P(Y=1)}$ and the imbalance ratio is $\rho=P(Y=1)/P(Y=0)=\pi/(1-\pi)$, an equivalent form for $\eta$ is $(\pi P)/(\pi P + (1-\pi)Q)$. A classification task is defined as a combination of prior probability $\pi$, loss function $\ell$ and $P,Q$ and denoted by $T=(\pi,P,Q;\ell)$. We formalize our major concern, the ultra-imbalance setting with respect to the classification task as below.

\begin{definition}
\label{def:Ultra-imbalance}
We say a classification 
task $T=(\pi,P,Q;\ell)$ is ultra-imbalanced if $\pi \rightarrow 0$.
\end{definition}

As mentioned before, our problem is set at \emph{population level} rather than sample level. The data is no longer separable even if the conditional class density is assumed to be sub-gaussian like in \cite{wang2021importance} and \cite{kini2021label}. \wrf{In this paper, we will be mostly concerned with how different loss functions behave under the UIC setup. To gain some insights, we first provide a rigorous analysis under a contrived example where the data are generated according to Gaussian mixture distribution.}

\subsection{A motivating case: analysis of Gaussian mixture}\label{sec: gaussian_mixture}

Let $[n]$ denote the set $\{1, \ldots, n\}$. Suppose the density of minority class is a mixture of $k_+$ Gaussian density with means $\{\mu_+^i\}_{i \in [k_+]}$, covariance matrix $\{\Sigma_+^i\}_{i \in [k_+]}$ and mixing weight $\{\pi_+^i\}_{i \in [k_+]}$, and the density of majority class is a mixture of $k_-$ Gaussian density with means $\{\mu_-^i\}_{i \in [k_-]}$, covariance matrix $\{\Sigma_-^i\}_{i \in [k_-]}$ and mixing weights $\{\pi_-^i\}_{i \in [k_-]}$. Mixing weight $\pi_-^i$ means the probability of a sample belonging to the $i$-th cluster in the majority class and $\pi_+^i$ is analogously defined. We have $\sum_i \pi_+^i=\sum_i \pi_-^i=1$. We mainly consider three different loss functions: 
\begin{description}
    \item[Square loss] $\ell^\text{mse}(y, \hat{y}) = (y - \hat{y})^2$.
    \item[(Proxy) cross entropy loss]  We will use the following \textbf{erf loss} function 
    \begin{align}
        \begin{aligned}
            \ell^\text{erf}(y,\hat{y}):=&y[u\Psi(u)-u+\Psi'(u)]\\
            +&(1-y)[-u\Psi(-u)+u-\Psi'(-u)],
        \end{aligned}
    \end{align}
   where $u=\log(\frac{1 - \hat{y}}{\hat{y}})$. It can be a proxy for the CE loss as it provides good approximation to CE, while enjoying close-form solutions when the underlying data generating distributions are Gaussian.
    \item[Alpha loss] Alpha loss \cite{Sypherd2022tunable} is a recently proposed loss function that unifies commonly used learning objectives like cross-entropy and exponential loss, with a hyperparameter $\alpha$ that controls the weight of poorly classified samples when $\alpha < 1$:
    \begin{align}\label{def: Alpha loss}
        \begin{aligned}
            \ell^\alpha(y,\hat{y}):=\frac{\alpha}{\alpha-1}&\left\lbrace\left[1-\hat{y}^{1-1/\alpha}\right]y\right. \\
            &\left.+\left[1-(1-\hat{y})^{1-1/\alpha}\right](1-y)\right\rbrace
        \end{aligned}
    \end{align}
\end{description}

The analysis will be based on the framework introduced in \cite{bach2006considering} that compares linear classifiers obtained by minimizing the above learning objectives.

To state our result, we introduce several additional definitions: We denote $\rho=\frac{\pi}{1-\pi}$ and call $f\sim g$ if $\frac{f(x)}{g(x)}\rightarrow{1}$ when $x\rightarrow 0$. We assume the linear classifier learned by loss $\ell$ is represented by $f(x)=\text{sign}(w_\ell x+b_\ell)$. Let $\Sigma_-=\sum_i\pi_-^i\Sigma_-^i+M_-(\text{diag}(\pi_-)-\pi_-\pi_-^T)M_-^T$, where $\mu_-=(\mu_-^1,...,\mu_-^{k_-})$, $M_-$ is matrix of the means of the clusters of minority class and $\mu_\pm=\sum_i \pi_\pm^i\mu_\pm^i$. We use $\text{diag}(v)$ to denote a diagonal matrix with diagnal elements being $v$. Let $\mu_+=\sum_i\pi_+^i\mu_+^i$,$\tilde{\mu}_-=\sum_i\xi_i\mu_-^i$ and $\tilde{\Sigma}_-=\sum_i\xi_i\Sigma_-^i$, with $\xi$ being the solution of the following convex program:
\begin{align*}
    \resizebox{\columnwidth}{!}{$
    \begin{aligned}
    &\min_{\xi} \sum_{i}\xi_i \log \xi_i - \sum_i \xi_i \left\{\log \pi_-^i-\theta(\xi)^T\mu_-^i+\frac{1}{2}\theta(\xi)^T\Sigma_-^i\theta(\xi)\right\} \\
    &s.t. \xi_i\geq 0,\quad \forall i, \sum \xi_i=1, \quad \theta(\xi) = (\sum_i \xi_i\Sigma_-^i)^{-1}\sum_{i}\xi_i\mu_-^i
    \end{aligned}
    $}
\end{align*}
Let $\check{\mu}_\pm =\sum_i\omega_\pm^i\mu_\pm^i$ and $\check{\Sigma}_\pm =\sum_i\omega_\pm^i\Sigma_\pm^i$. with $\omega_\pm \in \real_+^n$ being the solution to the following convex program:

\begin{align*}
    \resizebox{\columnwidth}{!}{$
    \begin{aligned}
      &\min_{\omega_+, \omega_-} \alpha\left[\sum_i \omega_-^i \left\{\log \frac{\omega_-^i}{\pi_-^i} + \theta(\omega)^T\mu_-^i - \frac{1}{2}\theta(\omega)^T\Sigma_{-}^i\theta(\omega)\right\}\right]\\
        &  +(1-\alpha)\left[\sum_i \omega_+^i \left\{\log \frac{\omega_-^i}{\pi_+^i} +\theta(\omega)^T\mu_+^i-\frac{1}{2}\theta(\omega)^T\Sigma_{+}^i\theta(\omega)\right\}\right] \\
        &s.t.\quad  \omega_-^i\geq 0, \omega_+^i\geq 0, \forall i, \quad \sum \omega_-^i=1,\quad \sum \omega_+^i=1,\\
        &\qquad \theta(\xi) = \left\{\alpha\sum_i \omega_-^i\Sigma_-^i+(1-\alpha)\sum_i \omega_+^i\Sigma_-^i\right\}^{-1}\sum_{i}\xi_i\mu_-^i
    \end{aligned}
    $}
\end{align*}

\begin{theorem}
\label{thm: square loss normal}
The following results characterizes the population risk minimizer regarding several losses under the UIC setup:\\
(i) \textbf{square loss}
\begin{align}
    w_\text{mse}\sim2\rho \Sigma_{-}^{-1}(\mu_+-\mu_-), b_\text{mse}\sim -1
\end{align}

(ii) \textbf{erf loss}:
\begin{align}
    \begin{aligned}
        &w_\text{erf}\sim (-2\log \rho)^{-1/2}\tilde{\Sigma}_-^{-1}(\mu_+-\tilde{\mu}_-), \\
        &b_\text{erf}\sim  -(-2\log \rho)^{1/2}
    \end{aligned}
\end{align}

(iii) \textbf{alpha loss}:
\begin{align}
    \begin{aligned}
        &w_{\alpha}\sim (\alpha\check{\Sigma}_-+(1-\alpha)\check{\Sigma}_+)^{-1}(\check{\mu}_+-\check{\mu}_-),\\
        &b_\alpha \sim \log \rho /\alpha
    \end{aligned}
\end{align}

(iv) \textbf{Optimality for a special case} If $k_+=k_-=1$, namely, the class conditional density are gaussian, the linear classifier learned by alpha loss with $\alpha = \frac{1}{2}$ has the optimal AUC among all linear classifiers. 

\end{theorem}

Theorem \ref{thm: square loss normal} implies that under UIC, even when infinite amount of samples are available, the linear classifiers obtained from three different losses put different emphasis on the minority class. In particular, alpha loss with lower $\alpha$ incorporates more covariance information from the minority class (reflected by the dependence on $\check{\Sigma}_+$). In contrast, the classifiers obtained from the square loss or cross entropy show no dependence over $\check{\Sigma}_+$. We will soon show simulated cases of normal mixture data, where focal loss, poly loss and vector scaling loss with constant multiplicative factor learns the totally same classifier as cross entropy. 

Furthermore, the alpha loss is provably optimal regarding AUC when instantiated as the exponential loss in a special case provided by (iv). The following simulated case shows the optimal choice of $\alpha$ vary with the setting of normal clusters.

\subsection{Numerical results from normal mixture models}\label{sec: normal_mixture}

We present a numerical case of normal mixture models with predictors of two dimensions. The imbalance ratio taken for simulation is $1:500$ and the size of minority class is 200. Both class are generated from two normal cluster. The $mean$ of two clusters in the majority class are set as $(2.0,2.0)^T,(2.0,-2.0)^T$ while the $mean$ of two clusters in the minority class are set as $(-2.0,2.0)^T,(-2.0,-2.0)^T$. The covariance of two clusters in the majority class are both identity matrix, while the covariance of the minority class are $\begin{psmallmatrix}0.5 & 0\\ 0 & 5\end{psmallmatrix}$
and $\begin{psmallmatrix} 5 & 0\\ 0 & 0.5\end{psmallmatrix}$.

\begin{figure}
\centering
\includegraphics[width=0.8\columnwidth]{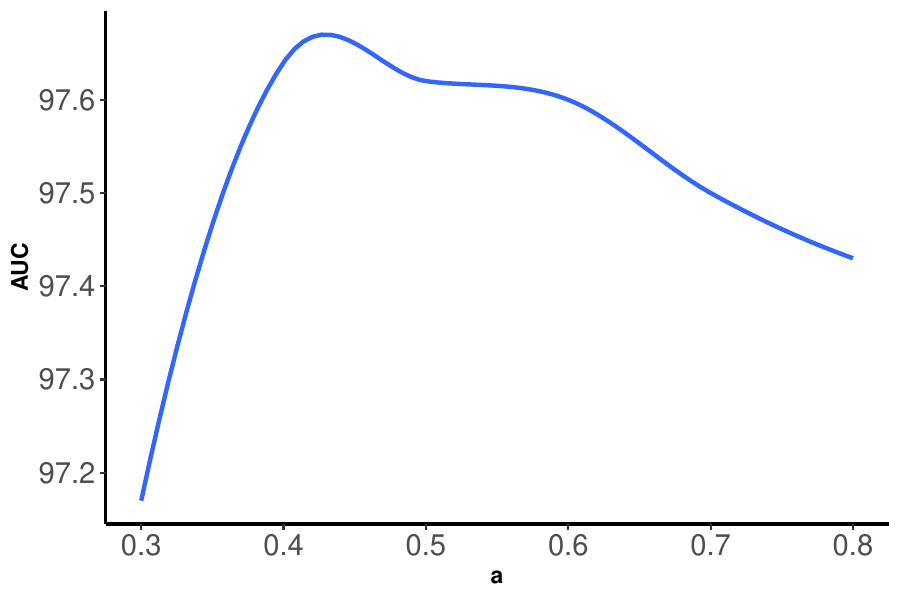}
\caption{X axis represents $\alpha$ used in learning linear classifier, Y axis represents the AUC value of the learned classifier in this case. The figure fits a smooth curve from results of different choice of $\alpha$ obtained by stochastic gradient descent.}
\label{fig: AUC}
\end{figure}

\begin{figure*}
    \centering
    \begin{subcaptionblock}{.4\textwidth}
        \centering
        \includegraphics[width=0.9\textwidth]{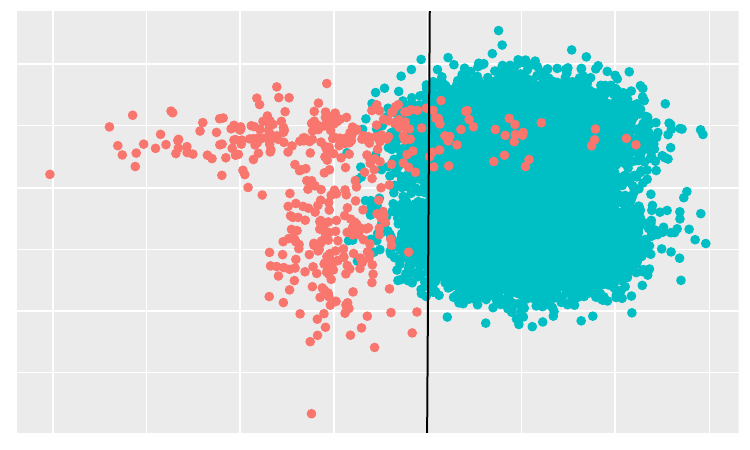}
        \caption{}
        \label{fig: simu_ce}
    \end{subcaptionblock}
    \begin{subcaptionblock}{.4\textwidth}
        \centering
        \includegraphics[width=0.9\textwidth]{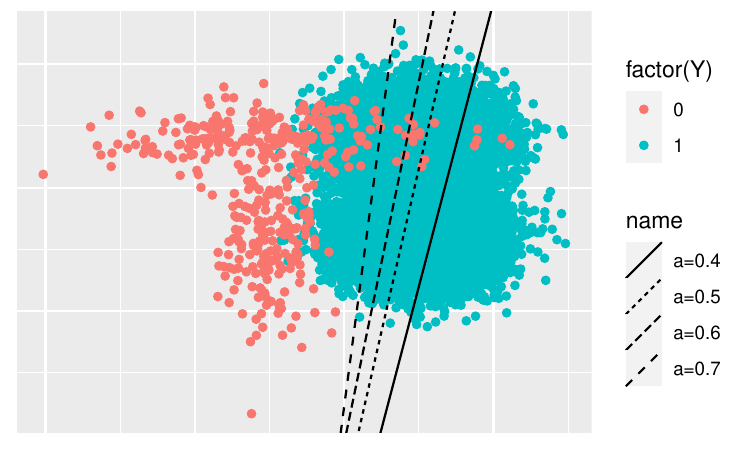}
        \caption{}
        \label{fig: simu_alpha}
    \end{subcaptionblock}
    \caption{Left: From the simulation, the linear classifier learned by Cross entropy, Focal loss, poly loss, and vector scaling loss is presented as the clear vertical line in the graph. So we do not show different lines. Right: Different type of lines represent linear classifier learned from alpha loss with different $\alpha$.}
\end{figure*}

Through figure \ref{fig: simu_ce}, we can clearly see focal loss and its variants do not incorporate the covariance information from the minority class, similarly as cross entropy. It means, though they are designed to reweight samples to tackle with imbalance, they do not make real change in learning classifier under UIC, at least in the case of normal mixtures. On the other hand, it is shown in figure \ref{fig: simu_alpha} that with $\alpha$ decreasing, the linear classifier learned by alpha loss tilts more to the minority class.

To present a further analysis of alpha loss in this case. Figure \ref{fig: AUC} records the variation of AUC of the learned linear classifier when $\alpha$ moves. The curve is fitted from 12 choices of $\alpha$. When $\alpha$ is around $0.4$, the AUC achieve its highest. It suggests that we may choose an appropriate $\alpha$ to optimize the AUC value.

While the results of theorem \ref{thm: square loss normal} is intriguing, it applies to a contrived example of Gaussian mixture. It is therefore of interest to develop principled methods for comparing different loss functions under UIC, with the underlying distribution allowed to be arbitrary.

\subsection{A framework for comparing loss functions}
\begin{table*}[t]\label{table: loss collection}
\caption{A summary of results under some loss functions}
\label{tab: f_function}
\begin{center}
\begin{small}
\begin{tabular}{l|l|l}
\toprule

Loss function  & Pointwise Risk & $f$-function \\
\toprule
Cross entropy &$-\eta\log(\hat{\eta})-(1-\eta)\log(1-\hat{\eta})$ & $-\pi \log(\pi) (1-t)$ \\
\midrule
Squared loss &$\hat{\eta}^2(1-\eta)+(\hat{\eta}-1)^2\eta$ & $\pi(1-t)$  \\
\midrule
Focal loss &$-\eta\log(\hat{\eta})(1-\hat{\eta})^\gamma-(1-\eta)\log(1-\hat{\eta})\hat{\eta}^\gamma$ &$-\frac{1}{\gamma+1}\pi\log(\pi)(1-t)$  \\
\midrule
Poly loss &$-\eta[\log(\hat{\eta})-\epsilon(1-\hat{\eta})]-(1-\eta)[\log(1-\hat{\eta})-\epsilon\hat{\eta}]$ &$-\pi \log(\pi) (1-t)$  \\
\midrule
VS loss &$\eta\log\left(1+(\frac{1-\hat{\eta}}{\hat{\eta}})^{\delta_1}\right)-(1-\eta)\log(1-\hat{\eta})$ &$-\delta_1\pi\log(\pi)(1-t)$ \\
\midrule
Alpha loss&$\frac{\alpha}{\alpha-1}\left\{\left[1-\hat{\eta}^{1-1/\alpha}\right]\eta +\left[1-(1-\hat{\eta})^{1-1/\alpha}\right](1-\eta)\right\}$ &$\frac{1}{1-\alpha}\pi^\alpha(1-t^\alpha)$\\
\bottomrule
\end{tabular}
\end{small}
\end{center}
\end{table*}

In this paper, we will base our comparison on the \emph{hardness} of the underlying classification task implied by different losses, which is closely related to the concept of statistical information \cite{degroot1962uncertainty}. To begin with, we define $\hat{\eta}:\mathcal{X}\rightarrow [0,1]$ as a class probability estimator. We next introduce some preliminary definitions, which are point-wise risk, point wise Bayes risk, and Bayes risk.

\begin{definition}
\label{def:point risk}
Point-wise risk at $x$ for $\hat{\eta}(x)$ and $\ell$ is the $\eta$-average of the point wise loss for $\hat{\eta}$, which is 
\begin{align}
    \begin{aligned}
        &L(\eta(x), \hat{\eta}(x)):=\mathbb{E}_{y \sim \eta}[\ell(y, \hat{\eta})]\\
        &\qquad =\ell(0, \hat{\eta}(x))(1-\eta(x))+\ell(1, \hat{\eta}(x)) \eta(x)
    \end{aligned}
\end{align}
\end{definition}
\vspace{-5pt}
\begin{definition}
\label{de:point wise bayes risk}
Pointwise Bayes risk at $x$ is the minimal achievable pointwise risk, which is defined as 
\begin{align}\label{def: Bayes risk}
\underline{L}(\eta):= \underset{\hat{\eta(x)}\in[0,1]}{\text{inf}}L(\eta(x),\hat{\eta}(x))
\end{align}
\end{definition}
\vspace{-5pt}
\begin{definition}
Bayes risk can be interpreted as the expectation of pointwise Bayes risk, which is 
\begin{align}\label{def: Bayes risk1}
\underline{\mathbb{L}}(\eta):=\int_{\mathcal{X}}\underline{L}(\eta(x))(\pi dP+(1-\pi)dQ)
\end{align}
\end{definition}
\vspace{-5pt}
\begin{definition}
\label{def: statisical information}
Statistical information is the difference of Bayes risk of the prior probability $P(Y=1)=\pi$ and the true conditional probability $\eta=P(Y=1|X=x)$:
\begin{align}\label{def: Bayes risk2}
\Delta\underline{\mathbb{L}}(\eta):=\underline{\mathbb{L}}(\pi)-\underline{\mathbb{L}}(\eta(x))
\end{align}
\end{definition}

The statistical information measures how much uncertainty is removed by knowing observation specific class probabilities $\eta$ rather than just the prior $\pi$. The smaller statistical information a classification has, the harder the task is. For example, the classification is impossible if $P=Q$ and the statistical information is 0 in that case, which also means knowing the prior $\pi$ is no more useful than knowing the true $\eta$ in classifying two classes. Statistical information serves as a useful criterion for comparison different loss functions. However, the precise form of statistical information depends on the underlying distributions (i.e., $P$ and $Q$) and is generally intractable. Therefore, we instead utilize the following alternative form of statistical information that is expressed as an $f$-divergence \cite{cover1999elements} between $P$ and $Q$, with the corresponding $f$ function depending on the prior probability $\pi$.
The statistical information can be alternatively expressed as the following $f$-divergence form
\begin{align}
    \Delta\underline{\mathbb{L}}(\eta)=\int f^{\pi}(dP/dQ) dQ,
\end{align}
with the corresponding $f$ function defined as
\begin{align}
    f^{\pi}(t) = \underline{L}(\pi) - (\pi t + 1 - \pi) \underline{L}\left(\dfrac{\pi t}{\pi t + 1 - \pi}\right).
\end{align}

$f$-function is often more tractable compared to the statistical information which shows the overall difficulty of a classification task. The following reformulation of $f$-function enables us to compare different loss functions under UIC. 
\begin{definition}[$f$-funtion under UIC]\label{def: f_uic}
    For any loss function $\ell$, a function $\tilde{f}: \mathbb{R} \times \mathbb{R} \mapsto \mathbb{R}$ is said to be an $f$-function under UIC, if $\tilde{f}$ satisfies $\lim_{\pi \rightarrow 0} \frac{\tilde{f}(t, \pi)}{f^\pi(t)} = 1$, with $f^\pi$ being the $f$-funtion of the corresponding statistical information induced by $\ell$.
\end{definition}

Hereafter we will refer to the function $\tilde{f}$ in definition \ref{def: f_uic} as the $f$-function of the underlying loss without further misunderstandings. When $\pi$ limits to zero, the statistical information will also limit to zero, which means the classification under ultra-imbalance is "infinitely" difficult. The proposed framework allows us to compare different loss functions by comparing the rates under which the $f$-function approaches zero. Next we compute the associating $f$-functions for two commonly used following loss functions: Cross entropy loss and square loss, as well as the following loss functions that were proposed to handle imbalanced learning problems:
\begin{description}
    \item[Focal loss \cite{lin2017focal}] with parameter $\gamma$ is defined as $\ell^\text{focal}(y,\hat{y})=-y\log(\hat{y})(1-\hat{y})^\gamma-(1-y)\log(1-\hat{y})\hat{y}^\gamma$.
    \item[Poly loss \cite{leng2022polyloss}] with parameter $\epsilon$ is defined as $\ell^\text{poly}(y,\hat{y})=-y\log(\hat{y})-(1-y)\log(1-\hat{y})+\epsilon \left[y(1-\hat{y})+(1-y)\hat{y}\right]$
    \item[Vector scaling loss \cite{kini2021label}] with parameter $\delta$ is defined as $ y\log\left(1+(\frac{1-\hat{y}}{\hat{y}})^{\delta}\right)-(1-y)\log(1-\hat{y}) $. 
    \footnote{The definition here is a slightly restricted version of the original proposal \cite{kini2021label}, we provide a discussion on the rationale of using this form in Appendix B. }
    \item[Alpha loss \cite{Sypherd2022tunable}] was defined in \eqref{def: Alpha loss}.
\end{description}

\begin{theorem}
\label{thm: standard loss}
We list the pointwise risk as well as the $f$-function of several useful loss functions in table \ref{tab: f_function}. 

\end{theorem}

According to theorem \ref{thm: standard loss}, on one hand, although focal loss, poly loss and VS loss have their different design of upweighting the minority class, the limiting behaviour of their corresponding $f$-function is almost the same (i.e., up to constants) under UIC. On the other hand, the $f$-function of alpha loss exhibits a slower decaying rate when $\alpha < 1$. It accords with the result in section \ref{sec: normal_mixture}, where focal loss and its variants do not make real difference out of cross entropy, while alpha loss give more emphasis on minority class when using a smaller $\alpha$. 

\subsection{Robustness improvements to the alpha loss}

\wrf{According to theorem \ref{thm: standard loss}, a smaller $\alpha$ configuration in the alpha loss achieves a stronger emphasis on the minority class  under UIC. However, this resistance comes at the cost of \textbf{worse robustness to outliers}. In particular, at $\alpha = 0.5$ the alpha loss is identical to the exponential loss \cite{Sypherd2022tunable} with its sensitivity to outliers been thoroughly discussed in previous works \cite{allende2007robust,rosset2003margin,ratsch1998regularizing}. To further analyze the robustness issue under general $\alpha$, we adopt the framework of influence analysis in robust statistics \cite{hampel1974influence}: Suppose the majority class and minority class are respectively sampled from $N(\mu_-,\Sigma_+)$ and $N(\mu_+,\Sigma_+)$, and we device a linear model for classification. For a specific point $z^* = (x^*, y^*)$ in the training sample, denote the \emph{influence} of upweighting $z^*$ evaluated with parameter $w$ as
\begin{align}
    \begin{aligned}
        &\mathcal{I}_\theta(z^*) = -H_{\theta}^{-1}\nabla_{\theta}\ell(y^*, h_w(x^*)), \\
        &H_\theta=\frac{1}{n}\sum_{i=1}^n\nabla_\theta^2\ell(y_i, h_\theta(x_i)),
    \end{aligned}
\end{align}
where $h_w(x)$ is the predictred label given $x$ with parameter $\theta$. In the case of linear model, $\theta = (w, b)$, and we have the following result:
\begin{theorem}\label{thm: influence}
    Under the linear model with Gaussian predictors and alpha loss, the influence of $z^* = (x^*, y^*)$ on parameters $\theta = (w, b)$ is 
    \begin{align}
        \mathcal{I}_\theta(z^*) = \dfrac{g(y^*,w^Tx^{*}+b)}{\sum_{i = 1}^n g(y_i,w^Tx_{i}+b)}\left(\frac{X^T X}{n}\right)^{-1}x^{*},
    \end{align}
    where $X$ denotes the sample feature matrix and the function $g$ is defined as
    \begin{align*}
        \begin{aligned}
            &g(y,w^T x+b) = 
            -y(1+e^{-y(w^Tx+b)})^{\frac{1}{\alpha - 2}}e^{-y(w^Tx+b)}\\
            &\qquad +(1-y)(1+e^{(1-y)(w^Tx+b)})^{\frac{1}{\alpha - 2}}e^{(1-y)(w^Tx+b)}
        \end{aligned}
    \end{align*}
\end{theorem}
According to theorem \ref{thm: influence}, for small $\alpha$ values, the sample fitted poorly has higher influence to the learned parameters, causing the model to exhibit poor robustness. To resolve this resistance-robustness trade-off, we propose the following improved version of alpha loss which we term \textbf{tunable boosting loss (TBL)}, where we directly incorporate penalization regarding observations with large influence:
\begin{align}\label{eqn: tbl}
\begin{aligned}
    \ell^\text{tbl}(y, \hat{\eta}):=&\frac{\alpha}{\alpha-1}\left[1-\hat{\eta}^{1-1/\alpha}\right]ye^{C(\hat{\eta}-1)}\\
    +&\frac{\alpha}{\alpha-1}\left[1-(1-\hat{\eta})^{1-1/\alpha}\right](1-y)e^{-C\hat{\eta}}
\end{aligned}
\end{align}
where $C$ is a hyperparameter to control how hard the influence is penalized. The bounded penalization terms $e^{C(\hat{\eta}-1)}$ and $e^{-C\hat{\eta}}$ are adaptions of influence penalization defined in \cite{Ratsch2001soft}. The limiting behavior of its f-function is unchanged for ultra-imbalance, with formal analysis deferred to Appendix C.

\begin{remark}
    So far we have devote all the effort to the case of binary classification. Extending our analysis to the multiclass case require suitably generalize the definition of statistical information \cite{duchi2018multiclass}. We will present a preliminary empirical exploration in section \ref{sec: multi-class} and leave theoretical discussions to future works.
\end{remark}
}

\begin{table*}[t]\label{table cat}
\vskip 0.15in
\begin{center}
\begin{small}
\input{binary_public}
\end{small}
\caption{Test best-1 accuracy ($\%$) and test auc ($\%$) on CIFAR-10, CIFAR-100 and Tiny ImageNet. We report each result as \result{\text{mean}}{\text{std}} obtained via $3$ trials. The best performance in mean is denoted in \textbf{bold}.}
\end{center}
\end{table*}

\section{Experiments}

\subsection{Experiment setups}
We present empirical evaluations with underlying classification task being treated as a UIC problem. We use two sources of datasets, with their summary statistics reported in appendix D: \\
\textbf{Image datasets} We conduct binary classification tasks on CIFAR-10, CIFAR-100 \cite{krizhevsky2009learning}, and Tiny ImageNet \cite{deng2009imagenet}. For each of the image datasets, we randomly select half of the categories as positives and the other half as negatives, in the main experimental comparison. We also utilize the CIFAR-10 deers and horses dataset in the ablation study. \\
\textbf{Fraud detection datasets} We use two industry-scale datasets collected from one of the world's leading online payment platforms. The task is a binary classification that aims at detecting fraudsters among regular users using a rich set of features. \\
\textbf{Training configurations} We use identical network architectures as in \cite{he2016deep} and \cite{arik2021tabnet}, with hyperparameter tuning procedures detailed in Appendix D. \\
\textbf{Baselines} We compare the classifier learned using our proposed TBL loss with those learned via the following objectives: cross entropy (CE) loss with logit adjustment, LDAM (label-distribution-aware) margin loss \cite{cao2019learning}, Focal loss with logit adjustment \cite{lin2017focal}, poly loss with logit adjustment \cite{leng2022polyloss}, VS (vector scaling) losses \cite{kini2021label}. All the hyperparameters involved in the baseline experiments are optimized using grid search, with the detailed configurations reported in appendix D.\\
\textbf{Evaluation metrics} For CIFAR-10, CIFAR-100 and Tiny ImageNet datasets, we use accuracy (ACC) and AUC as the evaluation metrics, since their test sets are balanced; For the two industrial datasets, we report AUC as well as two metrics that are crucial for evaluating models in the FRM domain: one-way partial AUC (\opauc) with an upper bound over false positive rate at $0.01$, and recall (\recall) at false positive rate $0.001$.  

\subsection{Results}

 \begin{figure}
 \centering
 \includegraphics[width=0.8\columnwidth]{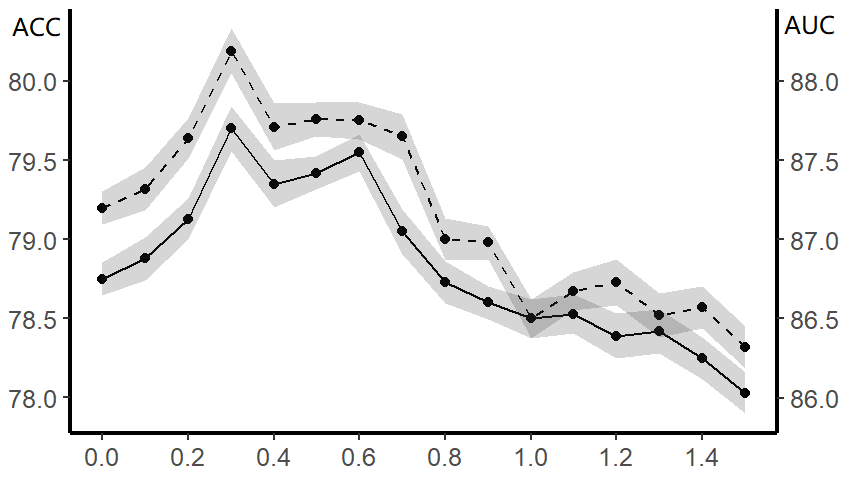}
 \caption{The line chart to reveal the effect of parameter $C$ with the confidence interval drawn. X axis represents the denoising parameter in tunable boosting loss, The solid line and the Y axis on the left represent the result of average accuracy. The dashed line and the secondary Y axis on the right represent the result of AUC. The shaded area reflects the standard error of result. See the text for interpretation.}
 \label{fig: varyingC}
 \end{figure}

    \begin{table*}[h]\label{table: fraud}
        \begin{center}
        \begin{small}
        \input{binary_ant.tex}
        \end{small}
        \end{center}
        \caption{Test AUC ($\%$), \opauc ($\%$) and \recall ($\%$) on two fraud detection datasets. We report each result as \result{\text{mean}}{\text{std}} obtained via $3$ trials. The best performance in mean is denoted in \textbf{bold}.}
    \end{table*}

\textbf{Image datasets} Tables 2 summarize the results of CIFAR-10, CIFAR-100 and Tiny ImageNet. 
Our proposed TBL loss consistently outperforms other methods in all scenarios, i.e. from the relatively easy task in CIFAR-10 to the extremely hard task in Tiny ImageNet. With the decrease of imbalance ratio $\rho$, the gain of TBL loss against CE loss also increases. 
As analyzed in section 2.3 and 2.4, all the chosen baselines do not significantly improve over the CE loss under UIC, with the TBL loss offering resistance against imbalance in the sense of a slower decaying rate. Therefore, the empirical results collaborate with our proposed theoretical framework. \\
\textbf{Fraud datasets} Tables 3 records the experiment results on the fraud-detection datasets. We observe from the experimental results that due to the strength high-quality features, all the methods exhibits competitive performance under the AUC metric, with a slight improvement achieved by TBL loss over the Fraud $1$ dataset. The difference in performance becomes more evident for the \opauc and \recall metric, under which TBL has the best overall performance, achieving dominating performance on the Fraud $2$ dataset. \\

\noindent
\textbf{The necessity of introducing $C$:} We conduct an ablation study to investigate
$C$ which promotes robustness. We expect a trade-off phenomenon to occur upon adjusting the values of $C$. The showcase is on the CIFAR-10 deers and horses dataset when $\rho=0.01$ and use ACC as well as AUC as the evaluation metric. Figure 4 records the variation of ACC and AUC when $C$ moves. It is clear to see when $C$ is around $0.3$, both the AUC and accuracy measure attain their maximums and are much better than $C=0$. It verifies the denoising design is beneficial to our tunable loss.

\subsection{Preliminary results on multi-class classification}\label{sec: multi-class}

Finally, we report an empirical investigation on the multi-class setup. We follow  \cite{cao2019learning} and consider the exponential-type imbalance and step-type imbalance with imbalance ratio $\rho\in \{0.1, 0.01\}$. We use the same set of baselines as in the binary experiments, with the implementation details provided in Appendix D. The results are reported in Table 4. The results demonstrate that the performance of TBL loss is comparable to or outperforming the most competitive baseline.

\begin{table}[ht]
\label{tab: multi-class}
\vskip 0.15in
\begin{center}
\begin{small}
\resizebox{\columnwidth}{!}{%
\input{multiclass.tex}
}
\end{small}
\end{center}
\caption{Test best-1 accuracy ($\%$) for CIFAR-10 dataset. We report each result as \result{\text{mean}}{\text{std}} obtained via $3$ trials. The best performance in mean is denoted in \textbf{bold}.}
\end{table}

\section{Conclusions and future works}

Motivated from the nature of modern financial risk management tasks, we formalize the concept of ultra imbalance classification (UIC) and reveal that loss functions can behave essentially different under UIC. We further borrow the idea of statistical information and develop a framework for comparing different loss functions under UIC. Finally, we propose a novel learning objective, Tunable Boosting Loss (TBL), which is provably resistant against data imbalance under UIC, as well as being empirically verified by extensive experimental studies on both public and industrial datasets.
In the future, we will further characterize the relationship between $f$-function under UIC and the linear classifier learned from more general distribution settings.

\bibliographystyle{named}
\bibliography{references}

\begin{appendix}
\onecolumn
\section*{Appendix A: Proof of Theorem 2}
(i), (ii) are the direct conclusions from Proposition 1 and 2 in \cite{bach2006considering}. Next we present the proof for (iii) first.

The alpha loss could be reformulated as a margin based loss function $\phi$, which is $\phi_{\alpha}(y,w^Tx+b)=e^{-y(w^Tx+b)}$.\cite{Sypherd2022tunable} In this case, the label of y is also reformulated into $1$ and $-1$. Label $1$ represents the majority class and the label $-1$ represents the minority class.

Set $u=\frac{1}{\alpha}-1$. 

When $x\sim N(\mu_-,\Sigma_-)$ then $w^Tx+b\sim N(w^T\mu_-+b,w^T\Sigma_- w)$, which is because $e^{-y(w^Tx+b)}$ will be very small.

Under ultra-imbalance, $$E\phi_{\alpha}(w^Tx+b)=\frac{\alpha}{\alpha-1}E[1-(1+e^{-y(w^Tx+b)})^{1/\alpha-1})]\approx e^{-(w^T\mu+b)+w^T\Sigma w/2}$$ 
When $x\sim N(\mu_+,\Sigma_+)$ then $w^Tx+b\sim N(w^T\mu_+ +b,w^T\Sigma_+ w)$.
Under ultra-imbalance, $$E\phi_{\alpha}(w^Tx+b)=\frac{\alpha}{\alpha-1}E[1-(1+e^{-y(w^Tx+b)})^{1/\alpha-1})]\approx 1/u e^{-u(w^T\mu_+ +b)+u^2w^T\Sigma_+ w/2}$$, which is because $e^{-y(w^Tx+b)}$ will be very large.

We denote $x_+^i\sim N(\mu_+^i, \Sigma_+^i), x_-^i\sim N(\mu_-^i, \Sigma_-^i)$. The derivative of training loss is zero, thus

$$\rho\sum_i \pi_+^i E\left[\frac{\partial \phi_\alpha(1,w^Tx_+^i+b)}{\partial w}\right]+\sum_i \pi_-^i E\left[\frac{\partial \phi_\alpha\left\{-1, w^Tx_-^i+b\right\}}{\partial w}\right]=0$$ and 
$$\rho\sum_i \pi_+^i E\left[\frac{\partial \phi_\alpha(1, w^Tx_+^i+b)}{\partial b}\right]+\sum_i \pi_-^i E\left[\frac{\partial \phi_\alpha\left\{-1, w^Tx_-^i+b\right\}}{\partial b}\right]=0$$ 
The calculation leads to
\begin{align}\label{key equation 1}
&\frac{1}{u}\rho\sum_i \pi_+^i E\left[e^{-u(w^T\mu_+^i+b)+u^2w^T\Sigma_+^i w/2}(u\Sigma_+^i w-\mu_+^i)\right] \nonumber\\ 
&+\sum_i \pi_-^i E\left[ e^{(w^T\mu_-^i+b)+w^T\Sigma_-^i w/2}(\Sigma_-^i w+\mu_-^i)\right]=0
\end{align}
and
\begin{align}\label{key equation 2}
-\frac{1}{u}\rho\sum_i \pi_+^i E\left[e^{-u(w^T\mu_+^i+b)+u^2w^T\Sigma_+^i w/2}\right]+\sum_i \pi_-^i E\left[ e^{(w^T\mu_-^i+b)+w^T\Sigma_-^i w/2}\right]=0
\end{align}

Divide (\ref{key equation 1}) by $\frac{1}{u}\rho\sum_i \pi_+^i E\left[\pi_+^ie^{-u(w^T\mu_+^i+b)+w^Tu^2\Sigma_+^i w/2}\right]=\sum_i \pi_-^i E\left[\pi_-^i e^{  (w^T\mu_-^i+b)+w^T\Sigma_-^i w/2}\right]$, we then have
$$\Sigma_i \tilde{\pi}_+^i\mu_+^i-\Sigma_i \tilde{\pi}_-^i\mu_-^i=(\Sigma_i \tilde{\pi}_-^i\Sigma_-^i+u\Sigma_i \tilde{\pi}_+^i\Sigma_+^i)w$$
where 
\begin{align}\label{final equation 1}
\tilde{\pi}_-^i=\frac{\pi_-^i \exp( w^T\mu_-^i+\frac{1}{2}w^T\Sigma_-^iw)}{\sum \pi_-^j \exp( w^T\mu_-^j+\frac{1}{2}w^T\Sigma_-^jw)}
\end{align}
and
\begin{align}\label{final equation 2}
\tilde{\pi}_+^i=\frac{\pi_+^i\exp(-u w^T\mu_+^i+\frac{1}{2}u^2w^T\Sigma_+^iw)}{\sum \pi_+^j \exp(-u w^T\mu_+^j+\frac{1}{2} u^2 w^T\Sigma_+^jw)}
\end{align}
we have
\begin{align}\label{final equation 2}
w=(u\sum_i\tilde{\pi}_+^i\Sigma_+^i+\sum_i\tilde{\pi}_-^i\Sigma_-^i)^{-1}(\sum_i\tilde{\pi}_+^i\mu_+^i-\sum_i\tilde{\pi}_-^i\mu_-^i)
\end{align}

which is 
$$w=\frac{1}{\alpha}(\alpha\sum_i\tilde{\pi}_+^i\Sigma_+^i+(1-\alpha)\sum_i\tilde{\pi}_-^i\Sigma_-^i)^{-1}(\sum_i\tilde{\pi}_+^i\mu_+^i-\sum_i\tilde{\pi}_-^i\mu_-^i)$$

The final part is to prove the solution of $\tilde{\pi}_-^i$, $\tilde{\pi}_+^i$ and $w$ is unique, which follows the path of proof of Proposition 2 in \cite{bach2006considering}. We assume $\xi=(\xi_-^1,...,\xi_{-}^{n_0},\xi_+^1,...,\xi_{+}^{n_1})$ and $\theta$ is the solution for $w$.

Let us define the following function defined on positive orthant $\left\{\xi,\xi_i>0,\forall i\right\}$.

\begin{align}
 H(\xi)=&\sum_i \xi_-^i \log \xi_-^i+\sum_i \xi_+^i\log \xi_+^i-\sum_i \xi_-^i \left\{\log \pi_{-}^i+\theta(\xi)^{\top} \mu_{-}^i+\frac{1}{2} \theta(\xi)^{\top} \Sigma_{-}^i \theta(\xi)\right\}\nonumber\\
 &-\sum_i \xi_+^i \left\{\log \pi_{+}^i-u\theta(\xi)^{\top} \mu_{+}^i+\frac{1}{2} u^2\theta(\xi)^{\top} \Sigma_{+}^i \theta(\xi)\right\}
 \end{align}

Calculation shows:
$$\frac{\partial\theta}{\partial\xi_+^k}=-(\sum_k\xi_-^k\Sigma_-^k+u\sum_k\xi_+^k\Sigma_+^k)^{-1}(\mu_+^k-u\Sigma_+^k\theta(\xi))$$
$$\frac{\partial\theta}{\partial\xi_-^k}=-(\sum_k\xi_-^k\Sigma_-^k+u\sum_k\xi_+^k\Sigma_+^k)^{-1}(\mu_-^k+\Sigma_-^k\theta(\xi))$$
$$\frac{\partial\left[\theta(\xi)^T\mu_-^i+\frac{1}{2}\theta({\xi})^T\Sigma_-^i\theta({\xi})\right]}{\partial \xi_-^j}=(\mu_-^i+\Sigma_-^i\theta(\xi))(\sum_k\xi_-^k\Sigma_-^k+u\sum_k\xi_+^k\Sigma_+^k)^{-1}(\mu_-^j+\Sigma_-^j\theta(\xi))$$
$$\frac{\partial\left[-\theta(\xi)^T\mu_+^i+\frac{1}{2}u\theta({\xi})^T\Sigma_+^i\theta({\xi})\right]}{\partial \xi_+^j}=-(\mu_+^i-u\Sigma_+^i\theta(\xi))(\sum_k\xi_-^k\Sigma_-^k+u\sum_k\xi_+^k\Sigma_+^k)^{-1}(\mu_+^j-u\Sigma_+^j\theta(\xi))$$
$$\frac{\partial H}{\partial \xi_+^i}=log\xi_+^i+1-\left[\log \pi_+^i-u\theta(\xi)^T\mu_+^i+\frac{1}{2}u^2\theta(\xi)^T\Sigma_+^i\theta(\xi)\right]$$
$$\frac{\partial H}{\partial \xi_-^i}=log\xi_-^i+1-\left[\log \pi_-^i-\theta(\xi)^T\mu_-^i+\frac{1}{2}\theta(\xi)^T\Sigma_-^i\theta(\xi)\right]$$
$$\frac{\partial^2 H}{\partial \xi_-^i \partial \xi_-^j}=\delta_{ij}\frac{1}{\xi_-^i}+(\mu_-^i+\Sigma_-^i\theta(\xi))(\sum_k\xi_-^k\Sigma_-^k+u\sum_k\xi_+^k\Sigma_+^k)^{-1}(\mu_-^j+\Sigma_-^j\theta(\xi))$$
$$\frac{\partial^2 H}{\partial \xi_+^i \partial \xi_+^j}=\delta_{ij}\frac{1}{\xi_+^i}+(\mu_+^i-u\Sigma_+^i\theta(\xi))(\sum_k\xi_-^k\Sigma_-^k+u\sum_k\xi_+^k\Sigma_+^k)^{-1}(\mu_+^j-u\Sigma_+^j\theta(\xi))$$
$$\frac{\partial^2 H}{\partial \xi_-^i \partial \xi_+^j}=\delta_{ij}\frac{1}{\xi_+^i}+(\mu_-^i-\Sigma_-^i\theta(\xi))(\sum_k\xi_-^k\Sigma_-^k+u\sum_k\xi_+^k\Sigma_+^k)^{-1}(\mu_+^j-u\Sigma_+^j\theta(\xi))$$
The last three equations show that the function H is strictly convex in the positive orthant. Thus minimizing $H(\xi)$ subject to $\sum_i \xi_i=1$ has an unique solution. Optimality conditions are derived by writing down the Lagrangian:
$$L(\xi,\alpha)=H(\xi)+\alpha(\sum_k\xi_+^k+\sum_k\xi_-^k-1)$$
which leads to the following optimality conditions:
$$\forall{i},\frac{\partial H}{\partial \xi_-^i}+\alpha=0,\frac{\partial H}{\partial \xi_+^i}+\alpha=0$$
$$\sum_i\xi_-^i=1,\sum_i\xi_+^i=1$$
These equations are equivalent to (\ref{final equation 1}), we have thus proved that the system defining $\theta$ and $\eta$ (Equation(\ref{final equation 1}),(4) and (\ref{final equation 2})) has a unique solution from the solution of the convex optimization problem:
\begin{center}
Minimize $H(\xi)$ with respect to $\xi$

such that $\xi_+^i\geq 0,\xi_-^i\geq 0,\forall{i}$

$\sum_i\xi_+^i=1,\sum_i\xi_-^i=1$
\end{center}
with
$$\theta(\xi)=\frac{1}{\alpha}(\alpha\sum_k\xi_+^k\Sigma_+^k-(1-\alpha)\sum_k\xi_-^k\Sigma_-^k)^{-1}(\sum_i\xi_+^i\mu_+^i-\sum_i\xi_-^i\mu_-^i)$$.

As for $b$, from (\ref{key equation 2}), we could solve that:
$$e^{\alpha b}=\rho C_0$$ namely $b=(\log \rho+\log C_0)/\alpha$ where $\log C_0$ is some constants compared to the diverging $\log \rho$. Directly we get $b=\ln{\rho}/\alpha$

\textbf{The proof of (iv):}
We assume two independent samples $x_+,x_-$ are respectively taken from $N(\mu_+,\Sigma_+)$ and $N(\mu_-,\Sigma_-)$

The AUC value in the case of two Gaussian cluster equals to
$$P(w^Tx_+\geq w^Tx_-)=\Psi\left\{\frac{w^T(\mu_+-\mu_-)}{\left[w^T(\Sigma_+ +\Sigma_-)w\right]^{1/2}}\right\}$$
where $\Psi(\cdot)$ is the cumulative distribution function of the standard normal distribution.

To maximize the using the conclusion of qudratic form, $w$ should be taken as $c(
\Sigma_++\Sigma_-)^{-1}(\mu_+-\mu_-)
$ where $c$ is an arbitrary nonzero constant.

The remaining part is to adapt the conclusion from (\ref{final equation 2}). When the number of normal cluster of each class is 1, the limiting solution of $(w,b)$ is exactly $(
\Sigma_++\Sigma_-)^{-1}(\mu_+-\mu_-)$. 

\section*{Appendix B: The justification of the modification of vector scaling loss}

Vector-scaling loss in \cite{kini2021label} is a combination of the multiplicative adjusting from CDTloss\cite{ye2020identifying} and the additive adjusting from logit adjustment \cite{menon2020long}. For a multiclass version, Vector-scaling loss is stated as
\begin{align}\label{def: VS loss}
\ell_{VS}(y,f_w(x))=\log(1+e^{\iota_y}\cdot e^{-\delta_y f_w(x)})
\end{align}
where $\iota_y=\tau \log P(Y=y)$ is the additive parameter, $\delta_y = P(Y=y)^\kappa$ is the multiplicative parameter and the vector $f_w(x)=(f_0(x),f_1(x))$ is the margin.

Plugging $e^\iota_y$ into loss function is equivalent to change the initial bias of the network to $(\iota_0,\iota_1)$ in the sense of training procedure. We consider this equivalence in a multiclass classification. Assume the prior probability for a sample belonging to class $i$ is $\pi_i$ and there are altogether k classes. After setting the initial bias according to the prior label distribution, which is $(\log \pi_1,..., \log \pi_k)$ 

If we represent the final output of the network with the initial bias subtracted by $f_y(x)$. The loss function could be formalized as
$$\ell(y,f(x))=-log\frac{e^{f_y(x)+\log \pi_y}}{\sum_{y'}e^{f_y(x)+\log \pi}}$$

The form is equivalent to the logit adjustment with the additive parameter being 1. See (10) in \cite{menon2020long}. It is noted that to calculate the probability from network output after changing the initial bias, one needs to calibrate the output by subtracting the initial bias.

The conclusions established on the population level are invariant of the initialization and non-margin-based loss functions can apply this easily. 

As for the multiplicative term of the minority class, it is set to be smaller than the majority class and all of the multiplicative terms are set to smaller than $1$. It means that if the margin of a sample is positive, then the sample will be upweighted more if its margin is closer to $0$. However, when data is ultra-imbalance, minority class samples are very likely to have negative margins, the multiplicative terms may have negative effect. Thus in our analysis, we choose the additive parameter to be zero, and the multiplicative parameter to be a constant smaller than $1$.

\section*{Appendix C: Proof of Theorem 8}

\textbf{Square loss:}
$$\underline{L}(\eta)=\eta(1-\eta)$$
The corresponding f-function is: $\pi(1-\pi)(1-\frac{t}{\pi t +1-\pi})\rightarrow \pi(1-t)$ when $\pi\rightarrow 0$

\textbf{Cross entropy:}
$$\underline{L}(\eta)=-\eta\log(\eta)-(1-\eta)\log(1-\eta)$$
The corresponding f-function is:
$-\pi\log(\pi)-(1-\pi)\log(1-\pi)+(\pi t)\log(\frac{\pi t}{\pi t +1-\pi})+(1-\pi)\log(\frac{1-\pi}{\pi t+1-\pi})$

According to the L'Hôpital's rule, when $\pi\rightarrow 0$ we have 
\begin{align}\label{lhoptial}
-(1-\pi)\log(1-\pi)=o\left(-\pi\log(\pi)\right), (1-\pi)\log(\frac{1-\pi}{\pi t+1-\pi})=o\left((\pi t)\log(\frac{\pi t}{\pi t +1-\pi})\right)
\end{align}

Thus we only need to consider the limit of $-\pi\log(\pi)+(\pi t)\log(\frac{\pi t}{\pi t +1-\pi})=-\pi\log(\pi)\left(1-t\frac{\log(\frac{\pi t}{\pi t +1-\pi})}{\log(\pi)}\right)$

Using L'Hôpital's rule again we have $\frac{\log(\frac{\pi t}{\pi t +1-\pi})}{log(\pi)}\rightarrow 1$ when $\pi\rightarrow 0$

Thus f-function of cross entropy limits to $-\pi\log(\pi)(1-t)$ when $\pi \rightarrow 0$

\textbf{Focal loss:} The point-wise risk of focal loss is 
$$L(\eta,\hat{\eta})=-\eta\log(\hat{\eta})(1-\hat{\eta})^\gamma-(1-\eta)\log(1-\hat{\eta})\hat{\eta}^\gamma$$
Solve $\frac{\partial L(\eta,\hat{\eta})}{\partial\hat{\eta}}=0$ and we have
$$-\eta\left\{\frac{(1-\hat{\eta})^{\gamma}}{\hat{\eta}}-\gamma\log(\hat{\eta})(1-\hat{\eta})^{\gamma-1}\right\}-(1-\eta)\left\{-\frac{\hat{\eta}^\gamma}{1-\hat{\eta}}+\gamma\hat{\eta}^{\gamma-1}\log(1-\hat{\eta})\right\}=0$$

It is easy to derive $\hat{\eta}\rightarrow 0$ as $\eta \rightarrow 0$ thus we omit the proof. Approximate $(1-\hat{\eta})$ and $(1-\eta)$ by 1, we get
\begin{align}\label{lemma focal}
\hat{\eta}^\gamma\left(\hat{\eta}-\gamma\log(1-\hat{\eta})\right)=\eta
\end{align}

From L'Hôpital's rule, we have $\frac{-\log(1-\eta)}{\eta}\rightarrow 1$ as $\eta\rightarrow 0$. Thus when $\eta\rightarrow 0$, $\hat{\eta}\approx (\frac{\eta}{1+\gamma})^{1/(\gamma+1)}$

The corresponding f-function could be derived similarly as proof of Theorem 3.8 (ii), which is
$-\frac{1}{\gamma+1}\pi\log \pi(1-t)$

\textbf{Poly loss:} is  
$$L(\eta,\hat{\eta})=-\eta\log(\hat{\eta})-(1-\eta)\log(1-\hat{\eta})+\epsilon(\eta(1-\hat{\eta})+(1-\eta)\hat{\eta})$$

Solve $\frac{\partial L(\eta,\hat{\eta})}{\partial\hat{\eta}}=0$ and we have
$$\hat{\eta}=\frac{2\eta}{\sqrt{\epsilon(1-2\eta)+1}+\sqrt{(\epsilon(1-2\eta)+1)^2-4\eta(1-2\eta)\epsilon}}$$
It approximates to $\eta$ when $\eta\rightarrow 0$.

The corresponding f-function approximates to $-\pi\log(\pi)+\pi t\log(\frac{\pi t}{\pi t+1-\pi})+2\epsilon\pi(1-t)$ with (\ref{lhoptial}).

Using L'Hôpital's rule again we have $\frac{\log(\frac{\pi t}{\pi t +1-\pi})}{log(\pi)}\rightarrow 1$ when $\pi\rightarrow 0$, and the f-function approximates to $-\pi\log(\pi)(1-t)$. 

\textbf{VS loss:} We restate the form of VS loss when $\delta_1<1$ and $\delta_{0}=1$ using the formula of logit $f_w(x)=\log(\frac{\eta}{1-\eta})$.

The point-wise risk of VS loss is:
\begin{align}\label{point risk vs}
L(\eta,\hat{\eta})=\eta\log\left(1+(\frac{1-\hat{\eta}}{\hat{\eta}})^{\delta_1}\right)-(1-\eta)\log(1-\hat{\eta})
\end{align}

Solve $\frac{\partial L(\eta,\hat{\eta})}{\partial\hat{\eta}}=0$ and we have
\begin{align}\label{lemma vs}
\delta_1\frac{(1-\hat{\eta})^{\delta}}{\hat{\eta}\left\{\hat{\eta}^{\delta_1}+(1-\hat{\eta})^{\delta_1}\right\}}=\frac{1-\eta}{\eta}
\end{align}

Plug in this back to (\ref{point risk vs}) we have
$$L(\eta,\hat{\eta})=\eta\log(\eta)-(\delta_1+1)\eta\log(\hat{\eta})-(1-\eta)\log(1-\hat{\eta})$$

The left side of (\ref{lemma vs}) is a monotone decreasing function of $\hat{\eta}$. Thus when $\delta_1<1$ and $\eta\rightarrow 0$, if $\hat{\eta}\leq \eta^b$ where $b>1$ is a constant, 

$$\delta_1\frac{(1-\hat{\eta})^{\delta_1}}{\hat{\eta}\left\{\hat{\eta}^{\delta_1}+(1-\hat{\eta})^{\delta_1}\right\}}\geq \delta_1\frac{1}{\eta^b\left\{(\frac{\eta}{1-\eta})^{b\delta_1}+1\right\}}\approx \delta_1\frac{1}{\eta^b}> \frac{1-\eta}{\eta}$$

It means $\hat{\eta}<\eta^b$ and $L(\eta,\hat{\eta})<-(b\delta_1+b-1)\eta\log(\eta)-(1-\eta)\log(1-\eta)$

On the other side, if $\hat{\eta}\leq \eta$, 
$$\delta_1\frac{(1-\hat{\eta})^{\delta_1}}{\hat{\eta}\left\{\hat{\eta}^{\delta_1}+(1-\hat{\eta})^{\delta_1}\right\}}\leq \delta_1\frac{1}{\eta\left\{(\frac{\eta}{1-\eta})^{\delta_1}+1\right\}}\approx \delta_1\frac{1}{\eta}< \frac{1-\eta}{\eta}$$

It means $\hat{\eta}>\eta$ and $L(\eta,\hat{\eta})>-\delta\eta\log(\eta)-(1-\eta)\log(1-\eta)$. 
Let $b\rightarrow 1^+$, we have $L(\eta,\hat{\eta})\approx\delta\eta\log(\eta)-(1-\eta)\log(1-\eta)$ and use the similar proof to Theorem 3.8 (ii) we derive its corresponding f-function is:
$$-\delta_1\pi\log(\pi)(1-t)$$

\textbf{Alpha loss: }According to (17) of \cite{Sypherd2022tunable}, the Bayes risk of tunable boosting loss is
$\underline{L}(\eta)=\frac{\alpha}{1-\alpha}(1-(\eta^\alpha+(1-\eta)^\alpha)^{1/\alpha})$

It is easy to prove $(1-(\eta^\alpha+(1-\eta)^\alpha)^{1/\alpha})$ approximates to $-\frac{1}{\alpha}\eta^\alpha$ when $\alpha$ is a rational number and $\eta\rightarrow 0$ by expanding $(\eta^\alpha+(1-\eta)^\alpha)^{1/\alpha})$. The approximation can be naturally extended to all real $a<1$. Thus the corresponding f-function approximates to $\frac{1}{1-\alpha}\pi^{\alpha}(1-t^{\alpha})$

\textbf{Justification of the f-function of (16):}

Calculate the derivative of $\tilde{\ell}^{\alpha}(\eta,\hat{\eta})$ to solve $\hat{\eta}$. We obtain the derivative which is:
\begin{align}\label{final analysis}
y\left\{\hat{\eta}^{-1/\alpha}e^{C(\hat{\eta}-1)}+Ce^{C(\hat{\eta}-1)}(1-\hat{\eta}^{1-1/\alpha})\right\}
+(1-y)\left\{-(1-\hat{\eta})^{-1/\alpha}e^{-C\hat{\eta}}-Ce^{-C\hat{\eta}}(1-\hat{\eta})^{1-1/\alpha}\right\}=0
\end{align}

When $\hat{\eta}\rightarrow 0$ under ultra-imbalance$, \hat{\eta}^{1-1/\alpha}$ is infinitely small compared to $\hat{\eta}^{-1/\alpha}$ and all of $e^{C(\hat{\eta}-1)}$ and $e^{-C\hat{\eta}}$ limits to a constant.

(\ref{final analysis}) can be reduced to
$$y\hat{\eta}^{-1/\alpha}e^{-C}=(Ce^{-C}+1)(1-y)(1-\hat{\eta})^{-1/\alpha}$$

The remaining proof follows the analysis of alpha loss.

\section*{Appendix D: Implementation details and other experiments}

\subsection*{A summary of datasets used in this paper.}

We summary the datasets and network architectures we used in Table 1. For more information on the Resnet or Tabnet, see \cite{he2016deep} and \cite{arik2021tabnet}.

 \begin{table}[H]\footnotesize\label{data feature}
 \label{sample-table}
 \vskip 0.15in
 \begin{center}
 \begin{sc}
 \begin{tabular}{c|c|c|c|c}
 \toprule

 Name & sample size & feature size& imbalance ratio & network \\

 \midrule
 Deer and horses in CIFAR-10 & 5000(majority class) & 32*32 & 0.002,0.01,0.05 & Resnet32 \\
 CIFAR-10 & 25000(majority class) & 32*32 & 0.1,0.05,0.01 & Resnet32\\
 CIFAR-100 & 25000 (majority class) & 32*32 & 0.1,0.05,0.01 & Resnet44\\
 Tiny ImageNet & 90000 (majority class) & 64*64 & 0.1,0.05,0.01 & Resnet56\\
 Fraud dataset 1 & 353310 & 71 & 0.01 & Tabnet\\
 Fraud dataset 2 & 940606 & 236 & 0.001 & Tabnet\\

 \bottomrule
 \end{tabular}
 \end{sc}
 \end{center}
 \vskip -0.1in
 \caption{A summary of datasets used in the papers}
 \vspace{30pt}
 \end{table}

\subsection*{Implementation details}\label{sec: implement}

For CIFAR-10, CIFAR-100, Tiny ImageNet data sets, we trained with SGD with a momentum value of 0.9 and use linear learning rate warm-up for first 10 epochs to reach the base learning rate, and a weight decay of $2.5*10^{-4}$.  We use a batch-size of 128. For data sets of CIFAR-10 and CIFAR-100, the base learning rate is set to 0.002, which is decayed by 0.1 at the 160th epoch and again at the 180th epoch. For dataset of Tiny ImageNet, the base learning rate is set to 0.01. For LDAM loss, the DRW traing rule proposed in \cite{cao2019learning} is adopted to give it an extra boost. The whole training lasts for 250 epochs.

For the first fraud detection dataset, we trained with AdamW with the base learning rate of $10^{-4}$ and the batch size is set to be 128. We use linear learning rate warm-up for first 15 epochs to reach the base learning rate.

For the second fraud detection dataset, we trained with AdamW with the base learning rate of $10^{-3}$ and the batch size is set to be 128. We use linear learning rate warm-up for first 15 epochs to reach the base learning rate

The ranges of searching the optimal parameters for all the datasets are the same and listed as follows,

\begin{itemize}
\item Focal loss: $\gamma\in\left\{1,1.5,2,2.5,3,3.5,4,4.5,5\right\}$;
\item Poly loss: $\epsilon\in\left\{-0.75,-0.5,-0.25,0.25,0.5,0.75,1,1.25,1.5\right\}$;
\item Vector Scaling loss : $\tau \in\left\{1,1.25,1.5,1.75,2\right\},\kappa \in \left\{0.1,0.15,0.2,0.25,0.3 \right\}$;
\item Tunable boosting loss: $\alpha \in \left\{0.7,0.75,0.8,0.85,0.9\right\}, C \in \left\{0.25,0.5,0.75,1\right\}$.
\end{itemize}

We record the optimal choice of parameters for all loss functions in Table\ref{optimal parameters1} and Table\ref{optimal parameters2}.

\begin{table}[H]
\vskip 0.15in
\scriptsize
\begin{center}
\begin{tabular}{lccc}
\toprule
Dataset & \multicolumn{3}{c}{Deer and Horses} \\
$\rho$  & 0.002  & 0.01 & 0.05  \\
\midrule
LDAM   & /  & /  & /  \\
Focal  & $\gamma=1$  & $\gamma=1$ & $\gamma=1$   \\
Poly   & $\epsilon=-0.75$  & $\epsilon=-0.75$ & $\epsilon=-0.5$ \\
VS     & $\tau=1.25,\gamma=0.1$  &$\tau=1.25,\gamma=0.1$ &$\tau=1.25,\gamma=0.1$ \\
TBL    & $\alpha=0.8,C=0.5$  & $\alpha=0.8,C=0.5$ & $\alpha=0.85,C=0.5$ \\
\toprule
Dataset & \multicolumn{3}{c}{CIFAR-10} \\
$\rho$  & 0.1  & 0.05 & 0.01  \\
\midrule
LDAM   & /  & /  & /  \\
Focal  & $\gamma=1$  & $\gamma=1$ & $\gamma=1$   \\
Poly   & $\epsilon=-0.5$  & $\epsilon=-0.5$ & $\epsilon=-0.5$ \\
VS     & $\tau=1.25,\gamma=0.2$  &$\tau=1.25,\gamma=0.2$ &$\tau=1.25,\gamma=0.2$ \\
TBL    & $\alpha=0.7,C=0.5$  & $\alpha=0.7,C=0.5$ & $\alpha=0.7,C=0.5$ \\
\toprule
Dataset & \multicolumn{3}{c}{CIFAR-100} \\
$\rho$  & 0.1  & 0.05 & 0.01  \\
\midrule
LDAM   & /  & /  & /  \\
Focal  & $\gamma=1$  & $\gamma=1$ & $\gamma=1$   \\
Poly   & $\epsilon=-0.5$  & $\epsilon=-0.5$ & $\epsilon=-0.5$ \\
VS     & $\tau=1.25,\gamma=0.2$  &$\tau=1.25,\gamma=0.2$ &$\tau=1.25,\gamma=0.2$ \\
TBL    & $\alpha=0.9,C=0.5$  & $\alpha=0.7,C=0.5$ & $\alpha=0.7,C=1.0$ \\
\toprule
Dataset & \multicolumn{3}{c}{Tiny ImageNet} \\
$\rho$  & 0.1  & 0.05 & 0.01  \\
\midrule
LDAM   & /  & /  & /  \\
Focal  & $\gamma=1$  & $\gamma=1$ & $\gamma=1$   \\
Poly   & $\epsilon=-0.5$  & $\epsilon=-0.5$ & $\epsilon=-0.5$ \\
VS     & $\tau=1.25,\gamma=0.2$  &$\tau=1.25,\gamma=0.2$ &$\tau=1.25,\gamma=0.2$ \\
TBL    & $\alpha=0.7,C=1.0$  & $\alpha=0.7,C=1.0$ & $\alpha=0.9,C=0.5$ \\
\bottomrule
\end{tabular}
\end{center}
\caption{optimal parameters choice for datasets of CIFAR-10, CIFAR-100, and Tiny ImageNet}
\label{optimal parameters1}
\end{table}
\vspace{30pt}

\begin{table}[H]
\vskip 0.15in
\footnotesize
\begin{center}
\begin{tabular}{lc|c}
\toprule
Dataset & Fraud detection data set 1 & Fraud detection data set 2\\
\midrule
LDAM & /  & /\\
Focal   & $\gamma=1.$  & $\gamma=1$  \\
Poly    & $\epsilon=1$  & $\epsilon=0.5$ \\
VS      & $\tau=1.5,\gamma=0$  &$\tau=1.25,\gamma=0$\\ 
TBL    & $\alpha=0.775,C=0.5$  & $\alpha=0.8,C=0.25$ \\
\bottomrule
\end{tabular}
\end{center}
\caption{optimal parameters choice for fraud detection datasets}
\label{optimal parameters2}
\end{table}

\vspace{30pt}

\subsection*{Comparison on the minority class accuracy}

\begin{figure}[H]
\vskip 0.2in
\begin{center}
\centerline{\includegraphics[width=0.9\columnwidth]{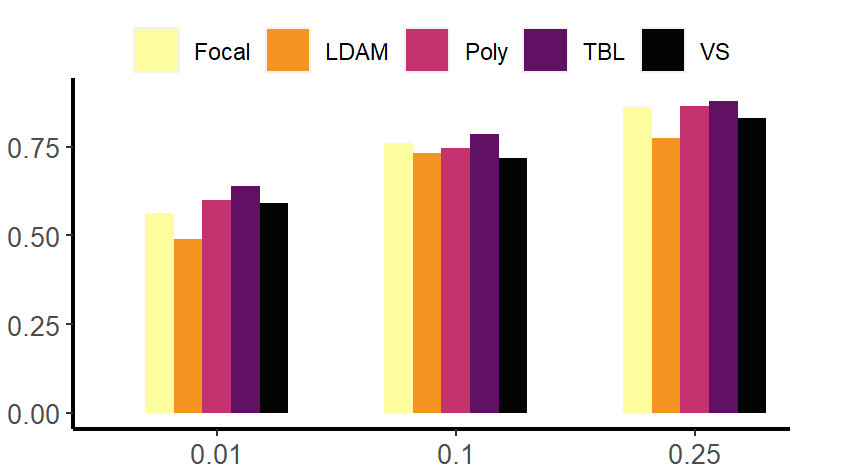}}
\caption{Bar chart on the minority class accuracy. Each bar represents the accuracy of minority class when the best test accuracy is achieved. The X axis represents the imbalance ratio $\rho$. See text for interpretation.}
\label{fig: worstcase}
\end{center}
\end{figure}

Our tunable boosting loss consistently improves the classification accuracy of the minority class over any other losses that upweight the minority class or samples difficult to identify. Figure \ref{fig: worstcase} demonstrates the minority class accuracy on the CIFAR-10 deers and horses dataset when taking $\rho=0.002,0.01,0.05$. TBL loss has $63.9\%$, $78.5\%$, $86.8\%$ accuracy of the minority class respectively, which is $3.8\%$, $2.9\%$, $0.8\%$ higher than the second-best result. The accuracy of the minority class of TBL loss is close to or even higher than the average accuracy.

\vspace{30pt}

\subsection*{Comparison on calibration ability}

The Brier Score is a strictly proper score function or strictly proper scoring rule that measures the accuracy of probabilistic predictions. The calibration ability is believed to be important in the classification accuracy.
\begin{table}[H]
\vskip 0.15in
\footnotesize
\begin{center}
\begin{tabular}{lccc}
\toprule
$\rho$  & 0.002  & 0.01 & 0.05 \ \\
\midrule
LDAM  & 0.215 & 0.183 & 0.130 \\
Focal  & 0.216  & 0.151& 0.115   \\
Poly  &  0.208  & 0.165 & 0.122 \\
VS    & 0.197  & 0.160 & 0.115 \\
TBL   & \textbf{0.190}  & \textbf{0.145} & \textbf{0.112} \\
\bottomrule
\end{tabular}
\end{center}
\caption{Brier score result for binary CIFAR-10 dataset of deers and horse}
\label{brier score}
\end{table}

\vspace{30pt}

We record the brier score result for the two binary CIFAR-10 dataset of deers and horses. It could be seen from Table \ref{brier score}, that TBL loss has a better ability of calibration over other losses. The reason is left to the future complement.

\end{appendix}

\end{document}

%% file: binary_public.tex
\begin{tabular}{c|c|cc|cc|cc}
    \toprule
    && \multicolumn{2}{|c|}{$\rho$ = 0.1} & \multicolumn{2}{|c|}{$\rho$ = 0.05} & \multicolumn{2}{|c}{$\rho$ = 0.01} \\
     Dataset&Method& ACC & AUC & ACC & AUC & ACC & AUC \\
    \midrule
    CIFAR-10&CE &\result{85.82}{0.201} &\result{93.90}{0.270} &\result{83.23}{0.379} &\result{91.80}{0.359} &\result{70.14}{1.505} &\result{85.41}{0.605} \\
    &Focal &\result{85.41}{0.523} &\result{93.50}{0.292} &\result{83.09}{0.573} &\result{91.43}{0.323} &\result{75.59}{1.103} &\result{85.10}{0.578} \\
    &LDAM &\result{85.72}{1.054} &\result{93.53}{0.361} &\result{79.38}{1.761} &\result{91.33}{0.361} &\result{75.52}{1.001} &\result{85.98}{0.534} \\
    &Poly &\result{86.30}{0.623} &\result{93.82}{0.396} &\result{83.72}{0.649} &\result{91.86}{0.639} &\result{77.58}{0.197} &\result{85.95}{0.382} \\
    &VS &\result{86.50}{0.078} &\result{94.08}{0.149} &\result{83.54}{0.119} &\result{91.70}{0.198} &\result{77.66}{0.501} &\result{85.93}{0.231} \\
    &\textbf{TBL(ours)} & \resultf{87.16}{0.428} & \resultf{94.65}{0.337}& \resultf{84.34}{0.185}& \resultf{92.75}{0.320}& \resultf{77.89}{0.163}& \resultf{86.41}{0.063}\\
    \midrule
    CIFAR-100&CE & \result{61.78}{0.701} & \result{68.98}{0.463} & \result{58.53}{1.377}&\result{65.31}{0.125}&\result{52.07}{0.630}	&\result{60.18}{0.270}\\
    &Focal & \result{62.58}{0.641}	&\result{68.69}{0.462}& \result{59.83}{0.547}&	\result{65.15}{0.393}&\result{55.40}{0.549}&	\result{59.77}{0.461}\\
    &LDAM & \result{57.74}{0.709}	&\result{66.89}{0.334}& \result{58.30}{0.231}&	\result{64.09}{0.211}&\result{55.21}{0.794}&	\result{59.31}{1.058}\\
    &Poly & \result{62.87}{0.202}	&\result{68.87}{0.146}&\result{60.14}{0.174}&	\result{65.45}{0.275}&\result{56.43}{0.675}&	\result{59.56}{1.031}\\
    &VS &\result{62.96}{0.332}&	\result{68.76}{0.458}&\result{60.16}{0.129}&	\result{65.29}{0.128}&\result{56.87}{0.451}	&\result{60.22}{0.413}\\
    &\textbf{TBL(ours)} & \resultf{63.43}{0.179}	&\resultf{69.40}{0.161}&\resultf{60.64}{0.285}&	\resultf{65.86}{0.349}& \resultf{57.30}{0.167}& \resultf{60.41}{0.210}\\
    \midrule
    Tiny ImageNet&CE & \result{51.21}{0.324}	&\result{56.00}{0.102} & \result{50.61}{0.242}&	\result{55.18}{0.113}&\result{50.22}{0.141}&	\result{54.04}{0.563}\\
    &Focal & \result{53.72}{0.136}&	\result{56.28}{0.270}& \result{53.34}{0.122}	&\result{55.15}{0.388}&\result{52.68}{0.156}	&\result{53.92}{0.281}\\
    &LDAM & \result{51.90}{0.359}	&\result{55.27}{0.172}& \result{51.41}{0.543}	&\result{54.94}{0.493}&\result{50.53}{0.167}	&\result{54.06}{0.250}\\
    &Poly & \result{53.53}{0.197}	&\result{55.94}{0.231}&\result{53.13}{0.313}	&\result{55.06}{0.176}&\result{52.34}{0.189}	&\result{53.95}{0.350}\\
    &VS &\result{53.81}{0.294}&	\result{55.70}{0.034}&\result{52.57}{0.073}&	\result{54.97}{0.543}&\result{52.27}{0.274}	&\result{53.37}{0.380}\\
    &\textbf{TBL(ours)} & \resultf{54.39}{0.459}	&\resultf{56.64}{0.564}&\resultf{53.45}{0.501}	&\resultf{55.45}{0.494}& \resultf{53.13}{0.488}&	\resultf{54.40}{0.654}\\
    \bottomrule
\end{tabular}

%% file: binary_ant.tex
\begin{tabular}{l|ccc|ccc}
    \toprule
    Dataset & \multicolumn{3}{c}{Fraud $1$} & \multicolumn{3}{c}{Fraud $2$} \\
    \midrule
    Criteria  & AUC  & opAUC & recall & AUC  & opAUC & recall \ \\
    \midrule
    LDAM & \result{96.13}{0.10} &\resultf{73.55}{0.25} &\resultf{32.76}{0.30} & \result{97.90}{0.02}  &\result{76.23}{0.18} & \result{38.44}{0.07} \\
    Focal   & \result{95.80}{0.14}  & \result{72.22}{0.23} & \result{29.91}{0.33} & \result{97.83}{0.10}  & \result{76.29}{0.21} & \result{38.37}{0.19}  \\
    Poly    & \result{96.14}{0.06} & \result{73.36}{0.21} & \result{31.82}{0.14} & \result{97.90}{0.02}  & \result{76.38}{0.16} & \result{38.57}{0.27} \\
    VS     &\result{96.12}{0.12}   & \result{73.26}{0.14} & \result{32.06}{0.23} & \result{97.90}{0.01} & \result{76.46}{0.16} & \result{38.76}{0.25} \\
    \textbf{TBL(ours)}    &\resultf{96.30}{0.08}  & \result{73.45}{0.18} & \result{32.45}{0.25} &\result{97.90}{0.01}  &\resultf{76.83}{0.43} &\resultf{39.14}{0.32} \\
    \bottomrule
\end{tabular}

%% file: multiclass.tex
\begin{tabular}{l|cc|cc}
\toprule

Imb. type & \multicolumn{2}{c}{Exp} & \multicolumn{2}{c}{Step}\\
\midrule 
 & $\rho = 0.1$ & $\rho = 0.01$ & $\rho = 0.1$ & $\rho = 0.01$\\
\midrule
LDAM &\result{87.64}{0.31}  & \result{77.03}{0.69}  & \result{87.82}{0.12} &\result{76.92}{0.37} \\
Focal  &\result{88.71}{0.08}  & \result{78.92}{0.12} & \result{88.64}{0.15}  & \result{78.76}{0.23}  \\
Poly   & \result{88.55}{0.13} & \result{78.18}{0.53} & \result{88.88}{0.19}  & \result{78.11}{0.42} \\
VS     & \result{88.72}{0.16}& \result{80.23}{0.54} & \result{88.93}{0.18} &\result{80.16}{0.21}\\ 
\textbf{TBL(ours)} & \resultf{89.08}{0.12} & \resultf{80.36}{0.32} & \resultf{89.16}{0.22} &\resultf{80.28}{0.28} \\
\bottomrule
\end{tabular}